\documentclass[letterpaper, 10 pt, journal, twoside]{IEEEtran} 
\usepackage[utf8]{inputenc}

\usepackage{units}
\usepackage{algorithm, algorithmic}
\usepackage{amsmath}
\usepackage{amssymb}
\usepackage{tabularx}
\usepackage{booktabs}
\usepackage{cite} 
\usepackage{graphicx}
\usepackage{subcaption}
\usepackage{mwe}
\usepackage{hyperref}
\usepackage[svgnames]{xcolor}
\usepackage{multirow}
\usepackage{makecell}
\hypersetup{
 colorlinks = true,
 linkcolor=black,
 urlcolor=Blue}

\usepackage{amsthm} 

\usepackage[colorinlistoftodos]{todonotes}

\newcommand{\rulesep}{\unskip\ \vrule\ }

\renewcommand\vec[1]{\boldsymbol{#1}} 

\newcommand{\de}{\;\mathrm{d}} 

\newif\ifcorrectingmode
\correctingmodefalse
\usepackage[normalem]{ulem}
\ifcorrectingmode
	
	\newcommand{\deleted}[1]{\textcolor{red}{\ifmmode\text{\sout{\ensuremath{#1}}}\else\sout{#1}\fi}}
	\newcommand{\deletedequation}[2]{\textcolor{red}{\centerline{Removed equation (#1)}}}
\else
	
	\newcommand{\deleted}[1]{}
	\newcommand{\deletedequation}[2]{}
\fi

\usepackage[printonlyused]{acronym}
\newacro{DOF}{degree of freedom}
\newacroplural{DOF}{degrees of freedom}
\acrodefplural{DOF}[DOFs]{degrees of freedom}
\newacro{iLQR}{Iterative Linear-Quadratic Regulator}
\newacro{CT}{Control Toolbox}
\newacro{EOM}{equations of motion}
\newacro{OC}{Optimal Control}
\newacro{LQR}{linear-quadratic regulator}
\newacro{PD}{proportional derivative}
\newacro{MPC}{Model Predictive Control}
\newacro{LQ}{linear quadratic}
\newacro{LQOC}{Linear-Quadratic Optimal Control}
\newacro{TO}{Trajectory Optimization}
\newacro{DDP}{Differential Dynamic Programming}
\newacro{COM}{center of mass}
\newacro{COP}{center of pressure}
\newacro{NLP}{nonlinear program}
\newacro{MLP}{Multilayer Perceptron}
\newacro{SLQ}{Sequential Linear-Quadratic}
\newacro{HAA}{hip abduction adduction}
\newacro{AD}{automatic differentiation}
\newacro{HJB}{Hamilton–Jacobi–Bellman}
\newacro{BC}{Behavioral Cloning}
\newacro{IRL}{Inverse Reinforcement Learning}
\newacro{IL}{Imitation Learning}
\newacro{RL}{Reinforcement Learning}
\newacro{MEN}{mixture-of-experts network}
\newacro{MILE}{mixture of implicitly localized experts}
\newacro{MELE}{mixture of explicitly localized experts}
\newacro{DL}{Deep Learning}
\newacro{WBC}{whole-body controller}
\newacro{MDP}{Markov Decision Process}
\newacro{RNN}{Recurrent neural network}
\newacro{PPO}{Proximal Policy Optimization}

\def\TheTitle{Combining Learning-based Locomotion Policy with Model-based Manipulation for Legged Mobile Manipulators}

\title{\LARGE \bf \TheTitle} 
\author{Yuntao Ma, Farbod Farshidian, Takahiro Miki, Joonho Lee, Marco Hutter%
    \thanks{This work was supported by the Max Planck ETH Center for Learning Systems, the Swiss National Science Foundation (SNSF) through project 166232, 188596, the National Centre of Competence in Research Robotics (NCCR Robotics), and the European Union's Horizon 2020 (grant agreement No.852044). Moreover, this work has been conducted as part of ANYmal Research, a community to advance legged robotics.}%
    \thanks{All authors are with the Robotic Systems Lab, ETH Z\"u{}rich, Switzerland. Email: {\tt\footnotesize mayun@ethz.ch}}%
}

\begin{document}

\maketitle
%
\thispagestyle{empty}
\pagestyle{empty}
%
%
\begin{abstract}
Deep reinforcement learning produces robust locomotion policies for legged robots over challenging terrains. To date, few studies have leveraged model-based methods to combine these locomotion skills with the precise control of manipulators. Here, we incorporate external dynamics plans into learning-based locomotion policies for mobile manipulation. We train the base policy by applying a random wrench sequence on the robot base in simulation and add the noisified wrench sequence prediction to the policy observations. The policy then learns to counteract the partially-known future disturbance. The random wrench sequences are replaced with the wrench prediction generated with the dynamics plans from model predictive control to enable deployment. We show zero-shot adaptation for manipulators unseen during training. On the hardware, we demonstrate stable locomotion of legged robots with the prediction of the external wrench. 
\end{abstract}
%
%
%

%
%


\section{Introduction}\label{sec:introduction}


Legged robots have potential to traverse rough unstructured terrain that are unreachable by traditional wheeled or tracked robots, utilize human-oriented infrastructures such as stairs and doors, and conduct various tasks by carrying task-specific payloads in harsh environments instead of human.

Mounting an arm and allowing the robot to manipulate in complex environments can largely extend their potentials. However, controlling legged robots with arms is inherently a high-dimensional, contact-rich, and often non-smooth control problem, and it is still an open problem.
To realize these potentials, many different control methods have been proposed. There are mainly two complementary directions: model-based control approaches and learning-based or data-driven approaches.

Model-based control algorithms such as \ac{MPC} have demonstrated precise control of wheeled mobile manipulators\cite{gawel2019fully} and robustness to external disturbances\cite{giftthaler2017efficient}. However, such methods face challenges for hybrid systems. In particular, for the legged mobile manipulators, the possible contact configurations drastically expand the search space of the trajectories, making the computational cost for optimal planning significantly higher and hard to meet real-time requirements.
To this end, many prior works use template models to reduce the computational cost, such as using zero-moment-point criteria~\cite{abe2013dynamic} or centroidal models\cite{dai2014whole}. With reduced systems, online planning of the motion trajectory becomes feasible on flat ground~\cite{bellicoso2019alma}\cite{sleiman2021unified}. Unfortunately, however, applying such methods for robots operating over rough terrain is not straightforward. There can be various disturbances, such as unplanned contacts and slipping, which make even a simple task like online end-effector tracking while walking difficult.

Model-free \ac{RL} has recently shown promising results to learn legged locomotion~\cite{hwangbo2019learning}\cite{lee2020learning}\cite{tan2018sim}\cite{miki2021wild}. In contrast to the model-based methods, \ac{RL} policies do not suffer from higher planning cost due to contacts, and learn to tackle unexpected contacts and slippage from environment interactions. 
Besides, a policy can be robustified by applying controlled disturbances during training. Existing works apply random external force and torques in the form of impulses\cite{peng2018deepmimic}, constant external wrench within the same episode\cite{lee2020learning}, or adversarial disturbances \cite{pinto2017robust} to make the policy robust against unobservable external disturbances and model mismatches.

\begin{figure}[t]
    \centering
    \subfloat[]{\includegraphics[width=0.225\textwidth]{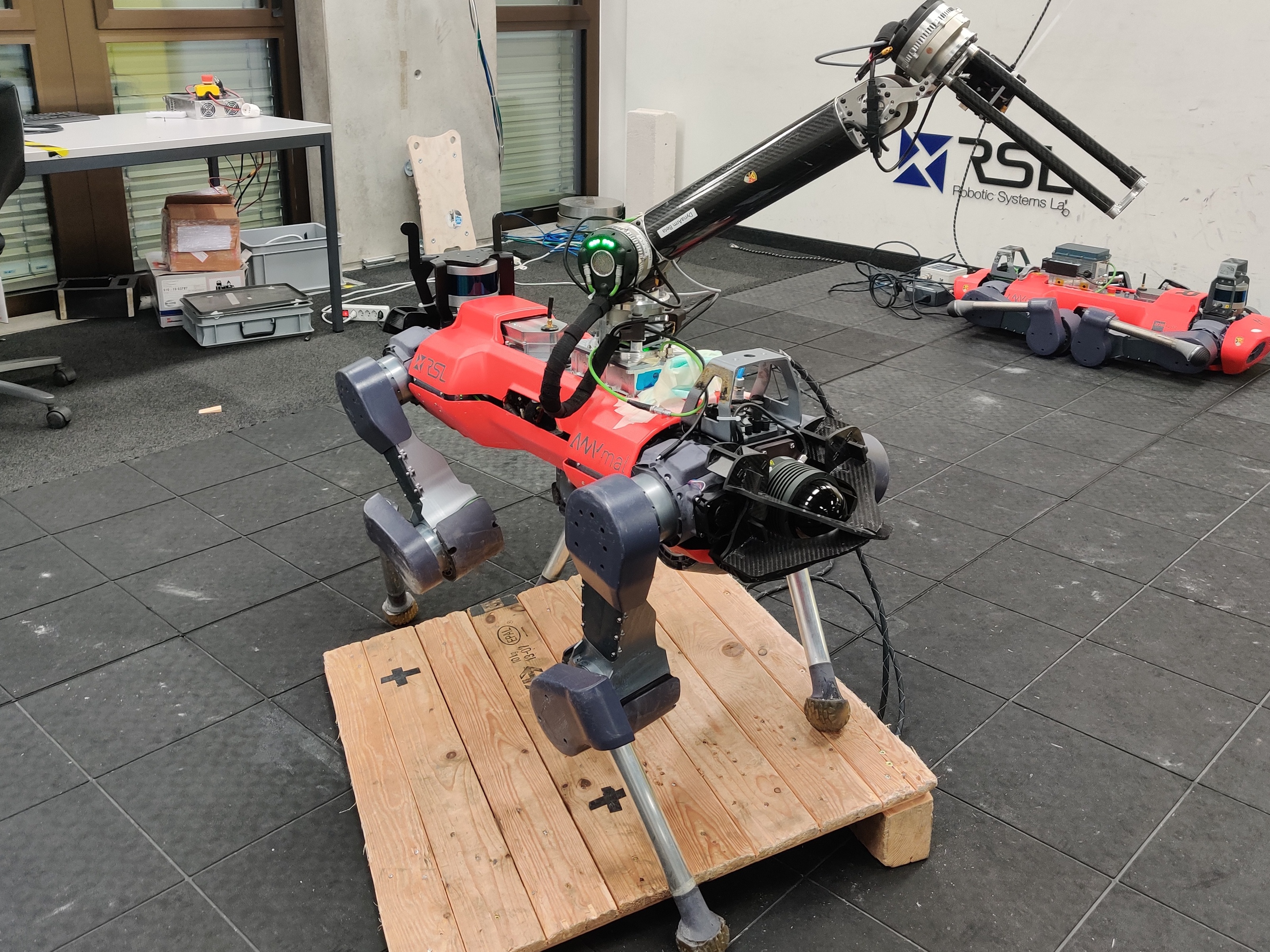}
    \vspace{-5pt}
    \label{subfig:bare_arm_board_cropped}}
    \vspace{1pt}
    \hfil
    \subfloat[]{\includegraphics[width=0.225\textwidth]{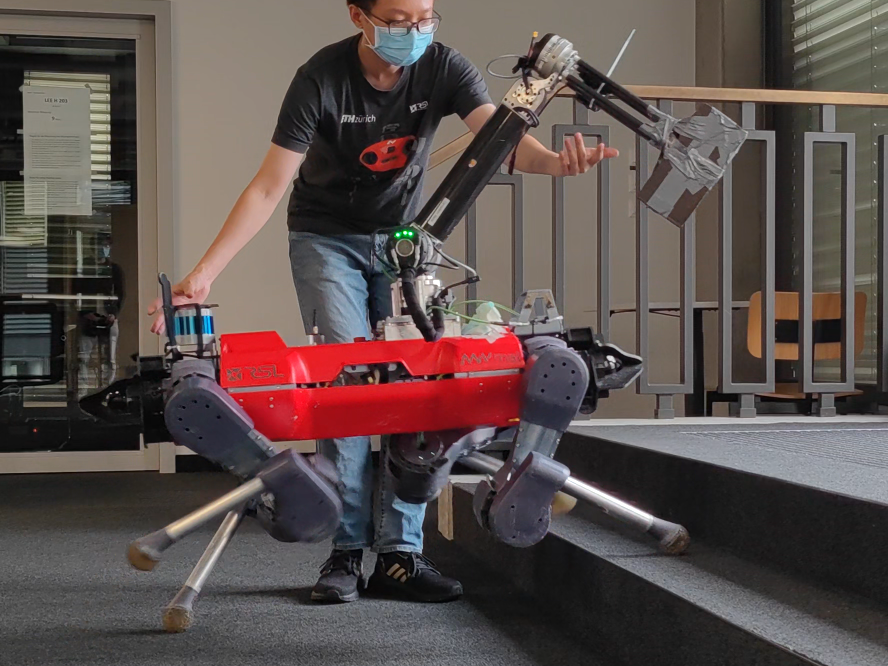}
    \vspace{-5pt}
    \label{subfig:brick_and_stair_cropped}}
    \vspace{1pt}
    \hfil
    \subfloat[]{\includegraphics[width=0.225\textwidth]{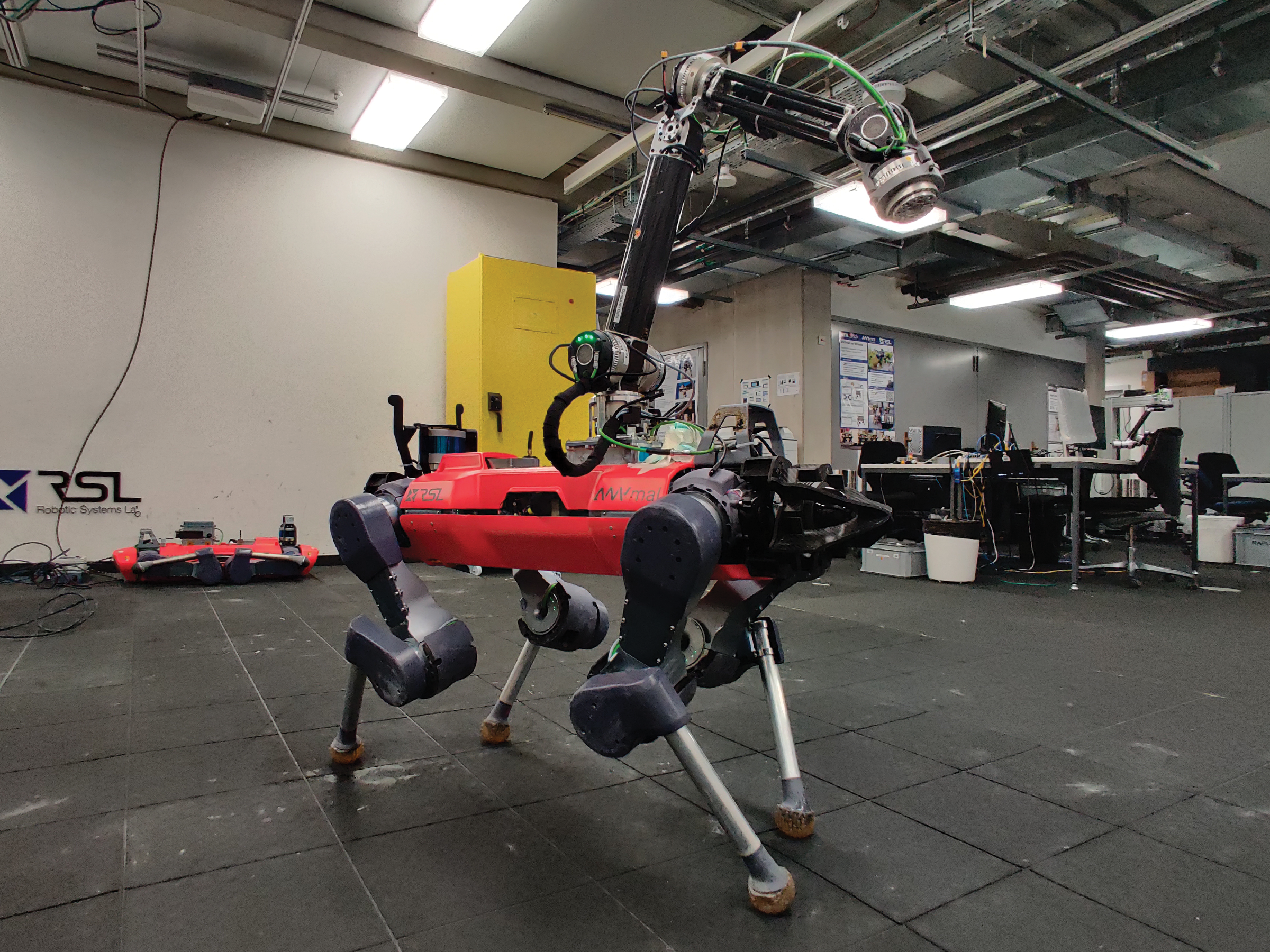}
    \vspace{-5pt}
    \label{subfig:wrist_flat_terrain_cropped}}
    \vspace{1pt}
    \hfil
    \subfloat[]{\includegraphics[width=0.225\textwidth]{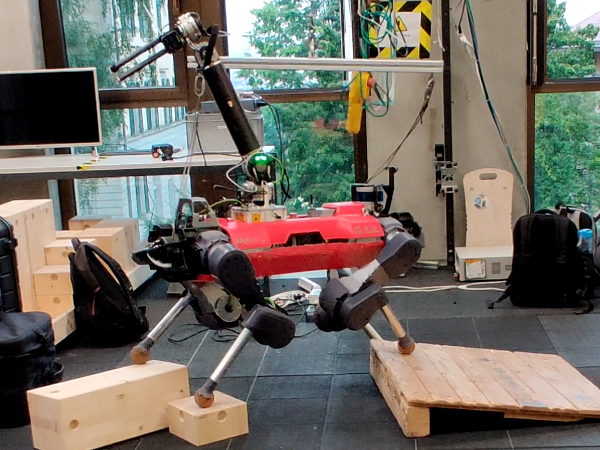}
    \vspace{-5pt}
    \label{subfig:alma_on_obstacles_cropped}}
    \vspace{1pt}
    \vspace{2pt}
    \caption{Experiments with ALMA. (\ref{subfig:bare_arm_board_cropped}-\ref{subfig:wrist_flat_terrain_cropped}) ALMA trotting with different manipulator configurations: no end-effector, fixed brick, a wrist weighing 1.8 kg. (\ref{subfig:alma_on_obstacles_cropped}) ALMA trotting on challenging terrain.}
    \vspace{-6pt}
    \label{fig:alma}
\end{figure}

Despite the simplicity and versatility of RL-based approaches, transferring a policy to diverse robot configurations \cite{finn2017model} and distinct tasks \cite{yin2019meta} still requires additional works. Usually, it requires re-training, which is time-consuming, and the policy can typically only be deployed on the robot that the policy is trained with, for the same task.  Some recent works explored combining \ac{RL} with model-based control and showed improved transferability. Prior works trained an \ac{RL} agent to output only the desired twist for the base~\cite{xie2021glide}, or the desired foothold on unstructured terrain~\cite{tsounis2020deepgait}. The model-based leg controllers then track these outputs.

Similarly, in this work, we split the legged mobile manipulator controller into two parts: the base and the arm controllers, which are separately constructed. For the base controller, we extend \cite{lee2020learning} and \cite{miki2021wild} by training a policy for the legged system that utilizes the external wrench {\color{Black}{(force and torque)}} observation to counteract disturbances. We further demonstrate that using wrench sequence prediction as observation improves the base \ac{RL} policy's performance over using only the current wrench. A proposed wrench sequence generator facilitates training of the policy without the arm controller. For deployment, we implement the arm controller with \ac{MPC}. It allows us to extract the wrench predictions for the base policy while controlling the arm precisely. The combined system demonstrates the capability of manipulation on rough terrain in hardware experiments. Additionally, since the training of the base policy is decoupled from the manipulator configurations, it can adapt to multiple arm setups without re-training. This zero-shot adaptation would not be feasible with a single controller of the entire system trained end-to-end with \ac{RL}.

Our main contributions are:
\begin{itemize}
\item A pipeline that combines model-based manipulator control and base locomotion policies trained with \ac{RL} for legged mobile manipulation.

\item Experiments showing the base locomotion policies having improved performance by using wrench predictions from \ac{MPC}. 

\item Zero-shot adaptation to multiple manipulator configurations on a legged robot platform for end-effector tracking.

\item Validation of the combined control system on a robotic hardware. The legged mobile manipulator walks on rough terrain while performing manipulation tasks.
\end{itemize}

%
%
\section{Background}

\subsection{Model Predictive Control}\label{ssec:mpc}

\ac{MPC} solves the following \ac{OC} problem
\stepcounter{equation} 
\begin{subequations}
\begin{align}
& \underset{\vec{u}(\cdot)}{\text{minimize}}
& & \phi(\vec{x}(t_f)) + \int_{t_s}^{t_f} l(\vec{x}(t),\vec{u}(t),t) \de t, \label{eq:cost} \tag{1} \\
& \text{subject to} & & \vec{x}(t_s) = \vec{x}_s, \\
& & & \dot{\vec{x}} = \vec{f}(\vec{x},\vec{u},t), \\
& & & \vec{g}(\vec{x},\vec{u},t) = \vec{0}, \\
& & & \vec{h}(\vec{x},\vec{u},t) \geq \vec{0},
\end{align}
\label{eq:constraints}%
\end{subequations}
where $\vec{x}(t)$, $\vec{u}(t)$ are the states and control inputs at time $t$ respectively. \ac{MPC} finds the minimum-cost~\eqref{eq:cost} trajectory under the system dynamics $\vec{f}$, the equality constraints $\vec{g}$ and the inequality constraints $\vec{h}$. The output of \ac{MPC} is the nominal trajectory $\vec{x}(t)$ and $\vec{u}(t)$. If the dynamics model is perfect and the terminal cost function $\phi(\vec{x}(t_f))$ captures the true terminal cost at $t_f$, the optimized state trajectories would be identical to the experienced state trajectories under the optimal input trajectories $\vec{u}(t)$ within the \ac{MPC} horizon. In practice, $\vec{u}(t_s)$ is then applied to control the robot in the receding horizon fashion to allow re-planning.

If the generalized positions, velocities and accelerations of the articulated system are available in the optimized state trajectory, an inverse dynamics approach such as the recursive Newton-Euler algorithm (RNEA) can provide the generalized forces consistent with these motions.

\subsection{Reinforcement Learning for Robots }\label{sec:background_rl}

To apply \ac{RL} for robotic control, a control problem is typically modeled as a discrete-time \ac{MDP}.  At each timestep, the \ac{RL} agent receives an observation $o_t\in\mathcal{O}$ from the environment and outputs an action $a_t\in\mathcal{A}$. The environment then updates with the action via the transition function $p(s_{t+1}|s_{t}, a_{t})$, evaluates a reward $r_t\in\mathcal{R}:\mathcal{A}\times\mathcal{S}\xrightarrow{} \mathbb{R}$ and generates a new observation for the agent. The agent may act according to a stochastic policy $\pi_{\theta}(a_t|o_t,o_{t-1},\dots)$. \ac{RL} algorithms train the agent by updating the parameters $\theta$ of the policies to maximize the cumulative discounted rewards $\mathbb{E}[\sum_{t=k}^{\infty}\gamma^t r_t]$ through interactions with the environments, where $k$ is the current timestep and $\gamma$ is the discount factor. In practice, the robot's policy observation usually consists of its state measurements, sensor readings such as a height scan, internal states to enforce desired behavior patterns, and possibly a history of the above.



%
%

\section{Method}

\begin{figure*}
    \centering
    \includegraphics[width=0.9\textwidth]{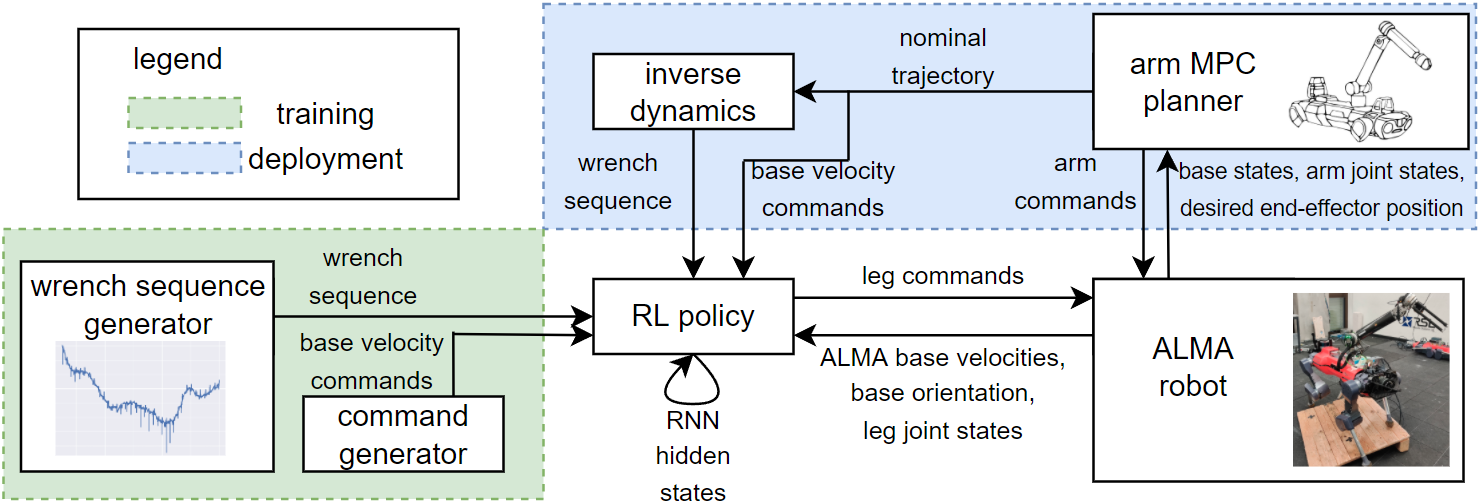}
    \vspace{6pt}
    \caption{Overview of the training and deployment pipeline. We train the \ac{RL} policy with a wrench sequence generator and replace it with the wrench plan from MPC for deployment.}
    \label{fig:pipeline}
    \vspace{-10pt}
\end{figure*}


\subsection{Overview}
{\color{Black}{Our method combines the precision and the planning capability of an MPC-based manipulation control and the robustness and the agility of an RL-based locomotion policy.}}
Fig.~\ref{fig:pipeline} shows the schematics of our method. We decouple the manipulator control from base policy training by modeling the disturbances from the manipulator as predictable external wrenches. The base control policy takes predicted wrench sequences as input and uses them to counteract the external wrench resulting from the arm. During the training, we simulate external wrench predictions using a wrench sequence generator without having the manipulator or solving \ac{MPC}. {\color{Black}{This drastically reduces the computational cost during the training process.}}

The floating-base arm is separately controlled by a whole-body MPC. When deploying on the robot, \ac{MPC} takes the user commands and produces nominal state and input trajectories of the floating-base manipulator, allowing for wrench sequence predictions. In this work, we deployed our controller on ALMA platform \cite{bellicoso2019alma}, which consists of the Dynaarm manipulator mounted on ANYmal base\cite{hutter2016anymal} and we use Raisim physics engine~\cite{raisim} to simulate the robot during locomotion training. 
 
The following sections describe the control problem for the manipulator system, the wrench sequence generator, and the structure of the \ac{RL} agent.

\subsection{MPC for the Manipulation System} \label{ssec:our_mpc}

Fig.~\ref{subfig:bare_arm_board_cropped}-\ref{subfig:wrist_flat_terrain_cropped} illustrate our different manipulator setups. For the $N$-dof arm system, we model the system as a manipulator on a floating base. 
\begin{align}
    \vec{x} &= [p_x, p_y, p_z, r_x,r_y,r_z, \theta_1, \dots \theta_N, \omega_1, \dots \omega_N]^T \label{eq:mpc_state} \\
    \vec{u} &= [v_x, v_y, \omega_z, \alpha_1, \dots \alpha_N]^T \label{eq:mpc_input}
\end{align}
The state, $\vec{x}$, consists of the base pose, manipulator joint positions $\theta$, and velocities $\omega$. The input, $\vec{u}$, is defined as the base linear velocity in $x, y$, angular velocity in $z$, and the joint accelerations $\alpha$. The system dynamics of the floating-base arm is defined in \eqref{eq:dynamics_part}. We choose {\color{Black}{a first-order system to model the base dynamics since we observe that the locomotion policy's transient behavior in tracking the command velocity resembles a first-order system closely. In contrast, we choose second-order dynamics for the manipulator's joints. Note we can directly calculate the joint torques through the inverse dynamics of the arm, and these two representations are essentially identical.}} 
\vspace{-5pt}
\begin{equation}
    \vec{f}_\text{base}(\vec{x}) = \begin{bmatrix}
    \dot{p}_x\\\dot{p}_y\\\dot{p}_z\\\dot{r_x}\\\dot{r_y}\\\dot{r_z}\\ \dot{\theta}_{1,\dots N} \\ \ddot{\theta}_{1,\dots N}
    \end{bmatrix}=\begin{bmatrix}
    v_x\\v_y\\-\kappa_1(p_z-h_\text{des})\\-\kappa_2 r_x \\-\kappa_3 r_y \\\omega_z\\ \omega_{1,\dots N} \\ \alpha_{1,\dots N}
    \end{bmatrix} \label{eq:dynamics_part}
\end{equation}
The system dynamics of the base, $\vec{f}_\text{base}$, is shown in \eqref{eq:dynamics_part}. We assume perfect base velocity tracking in $x$, $y$, and yaw direction. The base height, roll, and pitch have passive first-order dynamics that decay to their desired values, where the coefficients $\kappa$ are tuned from the {\color{Black}{trained student policy  described in Sec.~\ref{sssec:student}.}} 
This simplified model captures the based behavior that is trained to horizontally align the base at a desired height. 

The cost function for the end-effector tracking task consists of end-effector tracking cost, input cost, state deviation cost, which penalizes the arm joints for being away from their nominal values, and the joint velocity cost. In addition, we express joint position limits as inequality constraints. We solve the \ac{MPC} problem {\color{Black}{defined with (\ref{eq:mpc_state}), (\ref{eq:mpc_input}), (\ref{eq:dynamics_part}), and the above cost and constraints}} with \ac{SLQ} \cite{farshidian2017sequential} using OCS2~\cite{OCS2}.

Once the \ac{MPC} solution is found, the planned state-input trajectory is used to calculate the sequence of joint torques and the wrench sequence exerted on the base using an inverse dynamic formulation. In our implementation, we use Pinocchio \cite{pinocchioweb} for inverse dynamics. 

{\color{Black}{During deployment on the robot, dynamics mismatch between the systems and the simplified model for the base motion (\ref{eq:dynamics_part}) is inevitable. However, by running the MPC loop at a high rate (over 200 Hz), we ensure the state deviation from the planned trajectory is small.}}

\subsection{Wrench Sequence Generator} \label{ssec:mdp}
\begin{figure}
    \centering
    \subfloat[]{\hspace{-5pt}\vspace{-1pt}\includegraphics[width=0.22\textwidth]{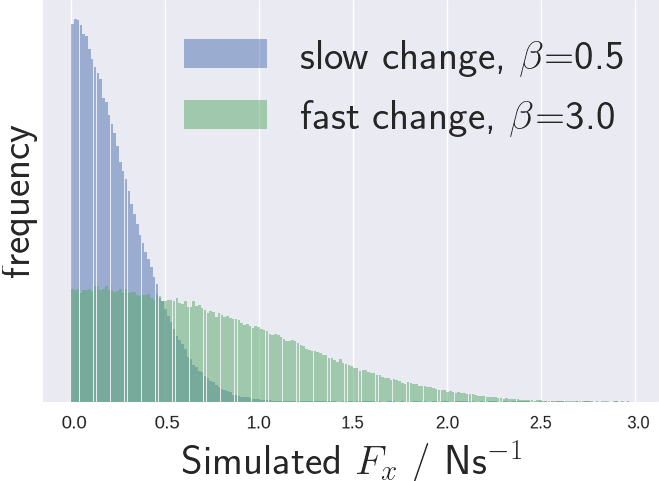}
    \vspace{-2pt}
    \label{subfig:rate_of_change}}
    \hfil
    \subfloat[]{\hspace{-5pt}\includegraphics[width=0.22\textwidth]{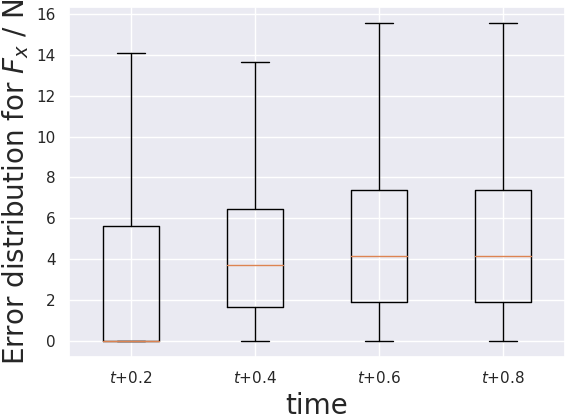}
    \vspace{-2pt}
    \label{subfig:pred_error}}
    \hfil
    \subfloat[]{\vspace{-3pt}\includegraphics[width=0.215\textwidth]{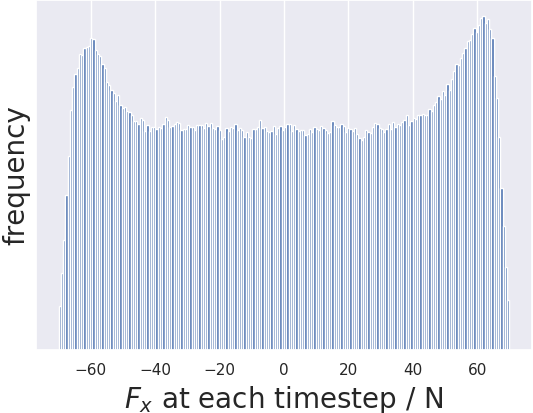}
    \vspace{-2pt}
    \label{subfig:wrench_distribution}}
    \hfil
    \subfloat[]{\includegraphics[width=0.235\textwidth]{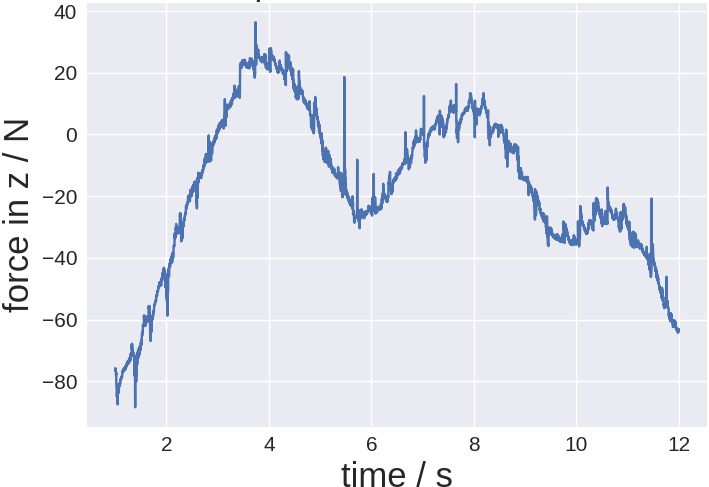}
    \vspace{-2pt}
    \label{subfig:example_wrench}}
    \vspace{2pt}
    \caption{Characteristics of the simulated wrench. (\ref{subfig:rate_of_change}) The rate of change of force in $x$ for different $\beta$. (\ref{subfig:pred_error}) Box plot of the prediction error for force in $x$ at time $t$ in the future. (\ref{subfig:wrench_distribution}) The distribution of force in $x$ that is applied on ANYmal. (\ref{subfig:example_wrench}) A sample of simulated force sequence in $z$.}
    \vspace{-1pt}
    \label{fig:wrench_char}
\end{figure}
During the \ac{RL} policy training, the wrench exerted by the manipulator is simulated by the wrench sequence generator.
Our wrench sequence generator consists of two models: Observed wrench and unobserved disturbances. 
{\color{Black}{\paragraph*{Observed wrench}
This component generalizes the smooth wrench from \ac{MPC} plan as the arm moves. We generate this component by with polynomials to capture its smoothness.}} We define a randomized quadratic function for each wrench dimension over the two seconds after the current time. 
\begin{equation}
    w_{t=0}, w_{t=1.0}, w_{t=2.0} \sim \text{Uniform}(w_\text{min}, w_\text{max}) \label{eq:wrench_generator_sample}
\end{equation}
\begin{equation}
    w_{t+2.0+dt} \sim \text{Uniform} (w_{t+2.0} - \beta w_\text{min}, w_{t+2.0} + \beta w_\text{max})\label{eq:wrench_generator_w3}
\end{equation}
At the start of each episode, we sample wrench $w_{t=0}$, $w_{t=1.0}$, $w_{t=2.0}$~\eqref{eq:wrench_generator_sample} and fit a quadratic function of $t$. {\color{Black}{Each sampled wrench $w$ is a six-dimensional vector representing the external force and torque in 3-D}}. For each subsequent timestep {\color{Black}{with $dt=\unit[0.02]{s}$}}, $w_{t+dt}$ and $w_{t+1.0+dt}$ are calculated from the the polynomial fitted in the previous timestep, and a random  $w_{t+2.0+dt}$ is sampled~\eqref{eq:wrench_generator_w3} and clipped between $w_\text{min}$ and $w_\text{max}$. $\beta$ randomizes the rate of change of the polynomial {\color{Black}{term}}, which is also sampled per episode. {\color{Black}{The range of $\beta$ values is tuned to produce smooth wrench that resemble those from the planned manipulator trajectories.}} $w_t$ is applied to the robot in world frame. For each dimension, we calculate the wrench at $[0.0, 0.2, 0.4, 0.6, 0.8]$ after the current time $t$ from the polynomial, and convert them to the robot base frame to feed them to the policy {\color{Black}{at each timestep during the training}}. The characteristics of the generated wrench is summarized in Fig.~\ref{fig:wrench_char}. 

\paragraph*{Unobserved disturbances} 
{\color{Black}{
In order to compute the induced arm wrench at the robot's base, we use inverse dynamics while ignoring the acceleration of the floating base. However, this assumption does not hold in practice due to the base motion, i.e., the cyclic pattern induced by the trotting gait. As a result, there is an unmodeled wrench component proportional to the inertia properties of the external loads: arm or other payloads. To make the policy robust against this component, we introduce a wrench disturbance induced by the base acceleration during the locomotion training.
}} At the start of the episode, two random vectors $\vec{v}_\text{force}$ and $\vec{v}_\text{torque}$ are sampled uniformly in spheres centered at the origin. At each simulation timestep, we multiply $\vec{v}_\text{force}$ and $\vec{v}_\text{torque}$ with the robot's base linear and angular acceleration respectively ($\vec{a}_\text{base}$ and $\vec{\alpha}_\text{base}$) and apply the results on the robot as wrenches. Furthermore, we add Gaussian noise to this unobserved disturbance to simulate the remaining disturbances, such as cogging in the manipulator drives.

At each timestep of the simulation, the sum of the two simulated external wrench is applied to the geometric center of ANYmal's base. The RL agent receives the observed wrench {\color{Black}{prediction sequence}} in the form shown in~\eqref{eq:wrench_observation}, where $w_t$ is the 6-d wrench queried at $t$ seconds after the current time.
%
\begin{equation}
    o_w = \begin{bmatrix}
    w_0, w_{0.2}, \dots, w_{0.8}, \text{commands}, v_\text{base},  \omega_\text{base} \end{bmatrix}^T \label{eq:wrench_observation}
\end{equation}
The predicted wrench sequence is expressed in the body frame at time $0$. The observed wrench and unobserved disturbance parameters are re-sampled at the start of each episode. A sample wrench sequence applied to the robot in the simulator is depicted in Fig.\ref{subfig:example_wrench}. 

\subsection{Base Policy Training}

\begin{figure*}
    \centering
    \subfloat[]{\includegraphics[width=0.39\textwidth]{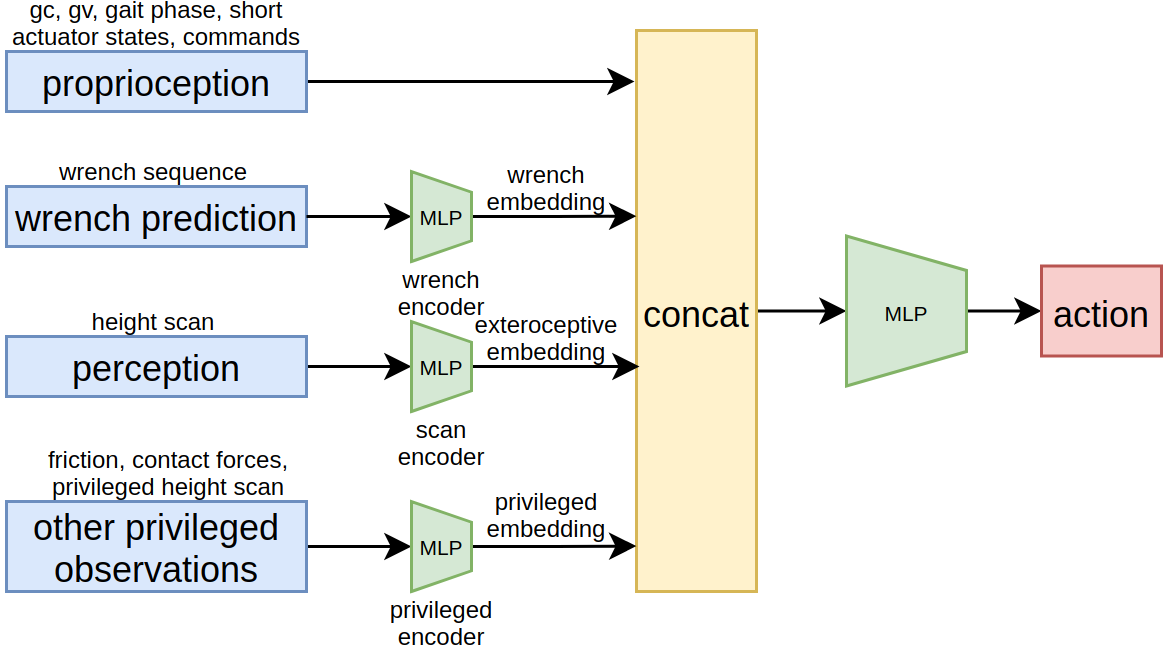}
    \label{fig:block_diagram_teacher}}
    \hfil
    \rulesep
    \subfloat[]{\includegraphics[width=0.58\textwidth]{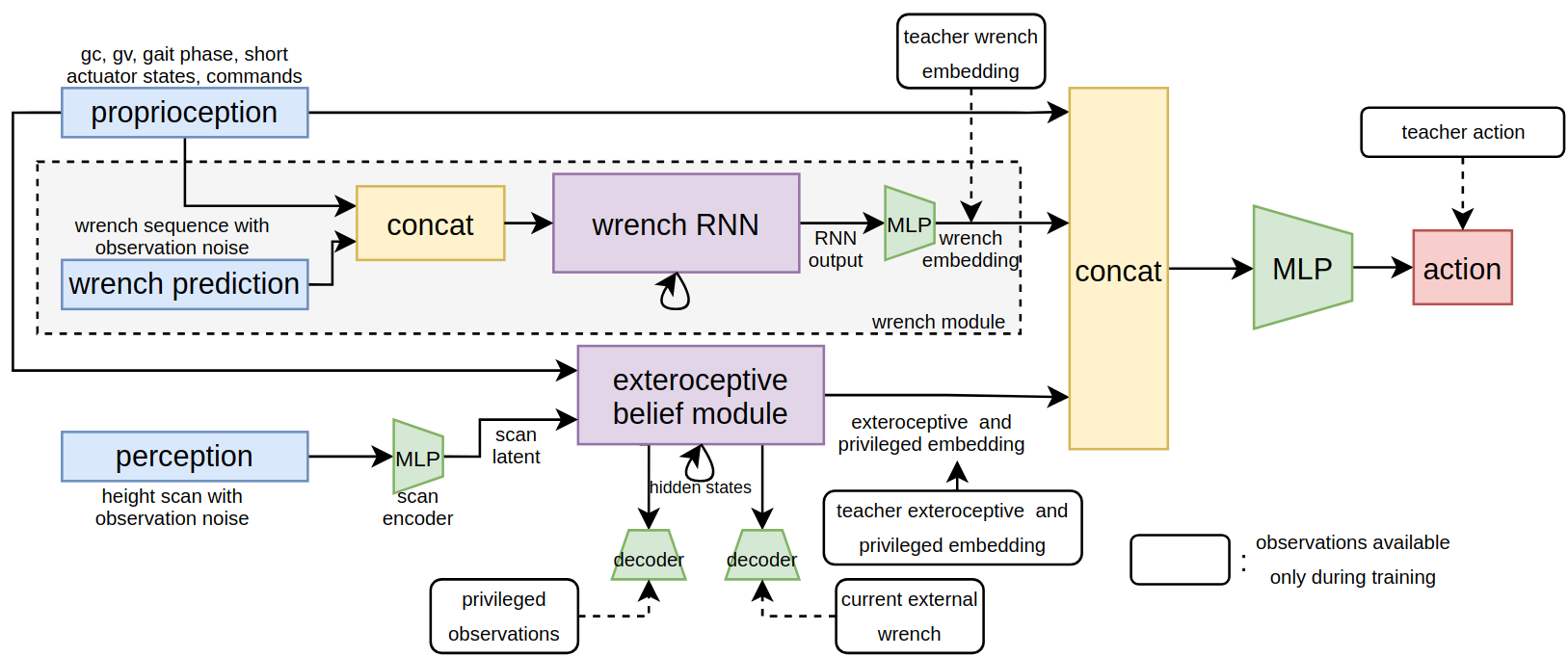}
    \label{fig:block_diagram_student}}
    \vspace{3pt}
    \caption{(\ref{fig:block_diagram_teacher}) Teacher policy structure. The teacher encodes the privileged observations with separate \ac{MLP}s. gc: generalized coordinate, gv: generalized velocity. (\ref{fig:block_diagram_student}) Student policy structure. The wrench module encodes the wrench prediction observation. The exteroceptive belief module is also trained to decode the external wrench from the proprioceptive history.}
    \vspace{-10pt}
    \label{fig:policy_structure}
\end{figure*}

We follow the privileged training method proposed in \cite{lee2020learning}. We first train a teacher policy (Fig.\ref{fig:block_diagram_teacher}) with privileged observation. The privileged observation includes accurate states of the simulation environment, such as ground friction coefficient or ground reaction forces, which are not directly observable in the real world. Then we train a student policy (Fig.\ref{fig:block_diagram_student}) which imitates the teacher's behavior without the privileged information. The student policy learns to extract meaningful information from the sequence of the non-privileged observations.

%

The teacher policy's wrench observation consists of the polynomial {\color{Black}{observed}} wrench, the base velocity commands, and the base twist (i.e., linear and angular velocities). For the student policy, we additionally noisify the wrench observations. The perturbed wrench observation is concatenated with the proprioceptive observation and passed to the wrench \ac{RNN} module, so that the policy may use the proprioceptive information to correct for the {\color{Black}{unobserved disturbances}}. {\color{Black}{In the case of quadrupedal robots, the teacher's policy converges to the trotting gait, and the student imitates this behavior.}} Readers can refer to {\color{Black}{\cite{lee2020learning}}} 
for the detailed format of the proprioceptive observation. {\color{Black}{The policies output the leg phases and a residual joint position. The phase here represents the progression of each foot in a cyclic stepping motion, and the residual joint position defines the offset joint angles of the joint-level PD controller. During the training, ANYmal is given a velocity command in $x$, $y$, and yaw direction for locomotion on rough terrain. }}

\subsubsection{Teacher}
The teacher agent is trained with \ac{PPO} \cite{schulman2017proximal} to maximize the expected cumulative trajectory rewards from the environment. The environment reward follows \cite{miki2021wild}, with a higher weight on the base orientation, the body velocity in $z$, and the base angular velocity in pitch and roll.

\subsubsection{Student} \label{sssec:student}
The student policy is distilled from the teacher policy by minimizing the loss function in~\eqref{eq:student_loss_big}, where $\mathcal{L}_\text{action}$ penalizes the difference between the student and teacher action, and $\mathcal{L}_\text{embedding}$ regresses the student wrench and exteroceptive embeddings towards those from the teacher.
\begin{align}
    \mathcal{L}_\text{total} &= \mathcal{L}_\text{action} + \mathcal{L}_\text{embedding} + \mathcal{L}_\text{decoder} \label{eq:student_loss_big}\\
    \mathcal{L}_\text{decoder} &= \mathcal{L}_\text{privileged} + \mathcal{L}_\text{scan} + \mathcal{L}_\text{w1} + \mathcal{L}_\text{w2} \label{eq:student_loss_decoder}
\end{align}
In~\eqref{eq:student_loss_decoder}, $\mathcal{L}_\text{w1}$ is the scaled mean-squared error between the decoded wrench and the current external wrench applied to the robot base. $\mathcal{L}_\text{w2}$ regresses the decoded unobserved wrench characteristics towards the magnitude of $\vec{v}_\text{force}$ and $\vec{v}_\text{torque}$ in Sec.~\ref{ssec:mdp}.
For manipulation tasks, the outputs from the decoder (e.g. load estimation) can be used for application purposes. We use two separate recurrent networks for the policy so that the decoder decodes the external wrench purely from proprioceptive and perceptive information without being interfered by the predicted wrench observations during the training. 


%
%

\section{Results}

This section shows the experimental results of our approach. The video recordings of selected experiments are available in Movie 1~\footnote{Movie 1: \href{https://youtu.be/OZ9Adh0PYbw}{https://youtu.be/OZ9Adh0PYbw}}.
\subsection{Teacher Policy Performance}
\begin{figure}
    \centering
    \includegraphics[width=0.3\textwidth]{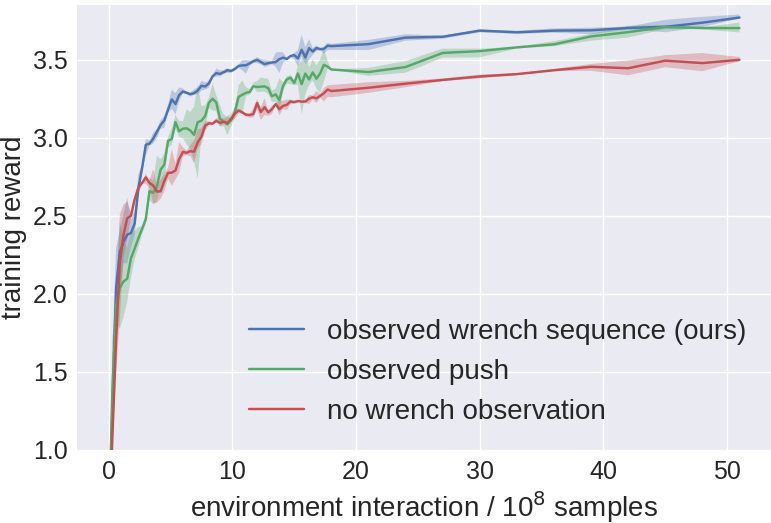}
        \vspace{3pt}
    \caption{Comparing training progress of the teacher policies.}
    \label{fig:compare_teacher}
\end{figure}

Fig.~\ref{fig:compare_teacher} compares the training progress of the teacher policy with and without the wrench {\color{Black}{prediction observations}} for {\color{Black}{five}} random seeds. It is clear from the curve that the external wrench prediction improves the teacher policy's learning progress and final performance. This shows that the teacher policy could indeed utilize the wrench prediction observations to counteract the disturbances. {\color{Black}{We additionally found that training with the smooth polynomial model as the predictable wrench results in a smoother learning progress than with intermittent pushes.}} We demonstrate with the other experiments in this section that the student policies can distill this ability.
%
\subsection{End-effector Tracking} \label{ssec:ee_experiment}
In this part, we deploy our proposed controller in simulation on ALMA {\color{Black}{to compare our method with the baselines for end-effector tracking}}. \ac{MPC} receives a moving end-effector target at each timestep and produces the motion plan that includes the base velocity commands for the base \ac{RL} policy, the arm joints trajectory, and predicted wrench induced at the base of the arm due to the arm motion. The performance of the base policy is measured with the mean {\color{Black}{absolute}} base tilt, the base angular velocity {\color{Black}{magnitude}} in pitch and roll, and the base linear velocity tracking error. We choose these evaluation metrics because they reflect the validity of the simplified base dynamics we described for \ac{MPC} in Sec.~\ref{ssec:our_mpc}. 
Two experiment settings are used to measure the performance of the combined control system. In both settings, we used Dynaarm with a 3 kg end-effector. To improve consistency, we set ALMA to always trot in these experiments.

\paragraph*{Experiment 1}
On rough terrain, a random reference point is selected in a square of $3\times 3  \text{ m}^2$, with height sampled from $\text{Uniform}(0.6, 1.1)$ m. The reference point is re-sampled every five seconds. Each experiment runs for half an hour in simulation time. This experiment assesses the performance of the policies with relatively static motion plans since the reference target is not re-sampled frequently. The base moves close to the max speed command of 0.5 ms$^{-1}$ when it is far away from the reference target.

\paragraph*{Experiment 2}
A reference point moves according to~\eqref{eq:experiment_2}, where $\tau$ is the time since the start of the episode in seconds. The \ac{MPC} commands the base to go to the origin with zero yaw.
\begin{equation}
\vspace{-4pt}
    p = [
    0.5, 0.5 \times \text{sin}(\tau/0.6), 1.1]^T \label{eq:experiment_2}
\vspace{-2pt}
\end{equation}
The manipulator moves dynamically in this experiment as the end-effector oscillates in $y$. Compared to Experiment 1, the external wrench from the manipulator changes more quickly.

\subsubsection{Comparing Student Policy Structures}\label{sssec:compare_student}

\begin{figure}
    \centering
    \subfloat[]{\includegraphics[width=0.22\textwidth]{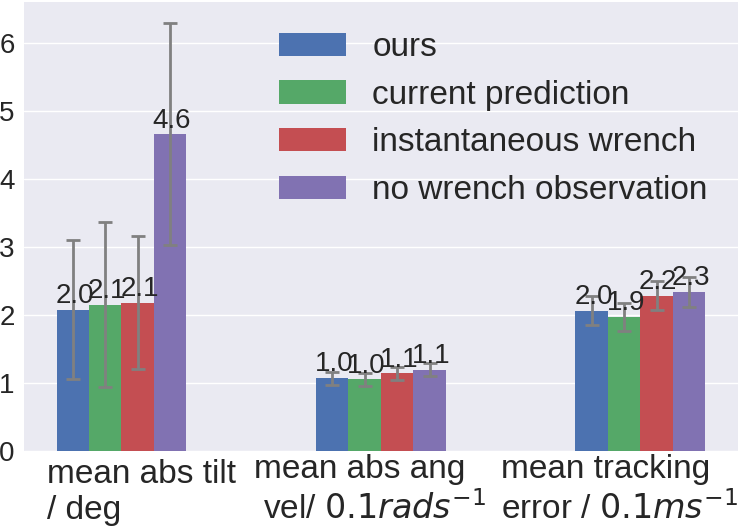}
    \vspace{-3pt}
    \label{subfig:compare_sim}}
    \vspace{3pt}
    \hfil
    \subfloat[]{\includegraphics[width=0.22\textwidth]{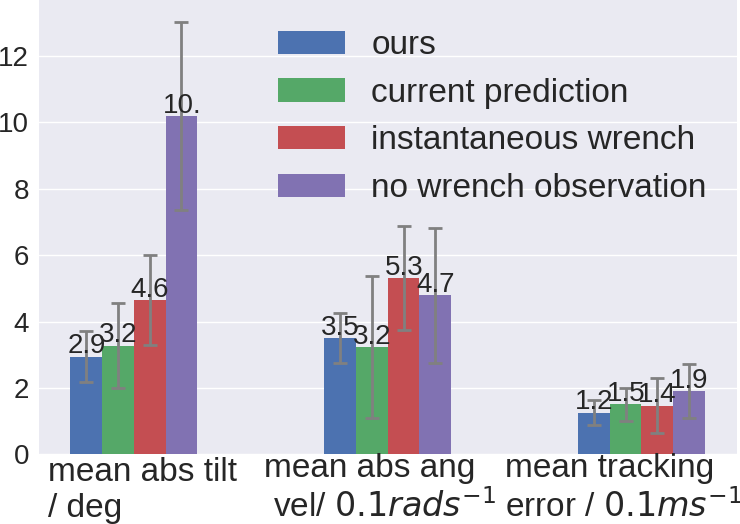}
    \vspace{-3pt}
    \label{subfig:left_right}}
    \vspace{3pt}
    \caption{Evaluation results of the tested student policy structures. (\ref{subfig:compare_sim}) Random end-effector targets. (\ref{subfig:left_right}) The end-effector target oscillates between the two sides of the robot.}
    \vspace{-1pt}
    \label{fig:compare_student}
\end{figure}

We ablate the students trained with the following settings, Policies in this experiments are each trained with {\color{Black}{five}} seeds using the same teacher policy for the same number of environment steps, with the same rewards.

\begin{itemize}
\item \emph{Ours}: The policy structure described in Sec.~\ref{sssec:student}.
\item \emph{Current prediction}: Two layer of \ac{MLP} is used instead of the \ac{RNN} wrench encoder. The policy cannot utilize the history of past wrench predictions.
\item \emph{Instantaneous wrench}: Instead of using the wrench prediction sequence of 1s, we only feed the wrench output calculated for the current timestep to the agent.
\item \emph{No wrench observation}: We remove the wrench module. The student policy only receives the proprioceptive and height scan observation.
\end{itemize}
Fig.~\ref{subfig:compare_sim} and  \ref{subfig:left_right} summarize the performance of different policy structures for the student policy. In general, it can be observed that the policies which utilize the external wrench predictions perform better. Close examination of these plots reveals a few interesting characteristics of our proposed controller.    

\emph{Explicit external wrench information is beneficial}: Even in Experiment 1 that
the manipulator has limited motion relative to the base, and the wrench is mainly from the weight of the manipulator, the external wrench improves the base tracking performance. Fig.~\ref{subfig:compare_sim} shows that all three policies with external wrench information have similar performance in base tilt and angular velocity, and they all outperform the policy with no external wrench observation. The mean tracking error assesses the base locomotion on rough terrain. The policy without wrench observation has a larger error because the base is more tilted, making it harder to walk.

\emph{Wrench prediction is beneficial during fast and dynamic motions}:
In Experiment 2, the manipulator has larger motions with respect to the base, making it more challenging for the base to balance. The wrench prediction contains information about the wrench trends, and the base policy could utilize this information to counteract the wrench. In the experiment, the policy that only observes the current wrench from the manipulator tilts to the side more and takes longer to recover than the policies that have access to the wrench prediction sequence. This is shown in Fig.~\ref{subfig:left_right}. In this experiment, the mean tracking error reflects the base's drift from the origin.

\emph{Note}: In both experiments, we did not apply any constant offset or scaling to the wrench prediction, so the policy that uses the current wrench prediction is expected to perform similarly to the one with the RNN encoder. This is also consistent with our experiment results.

The advantage of the RNN encoder over the policy that only encodes the current prediction depends on the wrench noise magnitude during the training. A larger noise makes the RNN policy rely on the proprioceptive observation and deteriorates the overall policy performance. The improvement is visible on the hardware, but for the sake of consistency, we do not include this comparison.

\subsection{Adaptation to Different Manipulator Configurations}

We assess the adaptability of the combined control system to various manipulator configurations by testing Experiment 2 with two additional settings in simulation.
\begin{itemize}
    \item Light arm: Dynaarm without any end-effector.
    \item Long arm: The forearm of Dynaarm is increased to 0.6~m, and the weight of forearm is adjusted accordingly. The end-effector is also removed in this setup. 
    \item Heavy arm: Dynaarm with the 3 kg end-effector. This is the same setting as Experiment 2 above.
\end{itemize}

\begin{figure}
    \centering
    \includegraphics[width=0.36\textwidth]{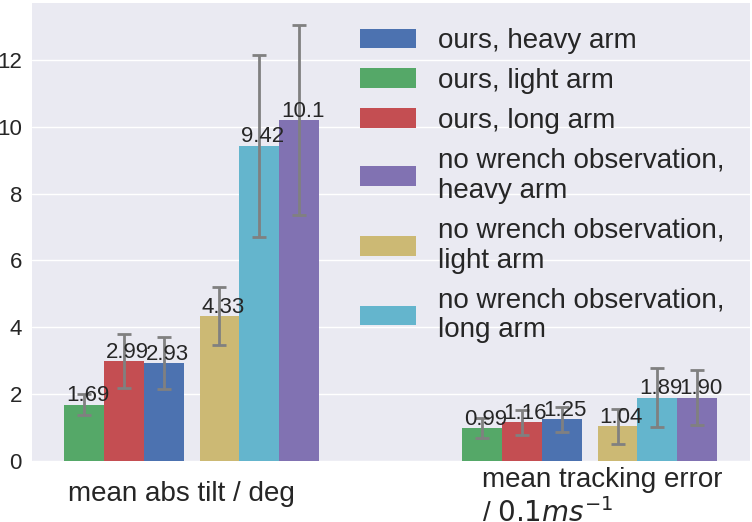}
    \vspace{4pt}
    \caption{Performance of the policy with different manipulator configurations.}
    \label{fig:different_configs}
\end{figure}

In this experiment, we compare our policy with the policy trained with no external wrench observation. When switching to manipulators that exert higher wrenches on the base, the policy that does not receive wrench observation tilts significantly more. In contrast, our policy is still able to keep the base close to horizontal (Fig.~\ref{fig:different_configs}). This experiment shows that our policy can adapt to various manipulator configurations with a much smaller performance drop than the policy with no wrench observation.

\subsection{End-effector Tracking under Base Disturbance}

\begin{figure}
    \centering
    \subfloat[]{\includegraphics[width=0.228\textwidth]{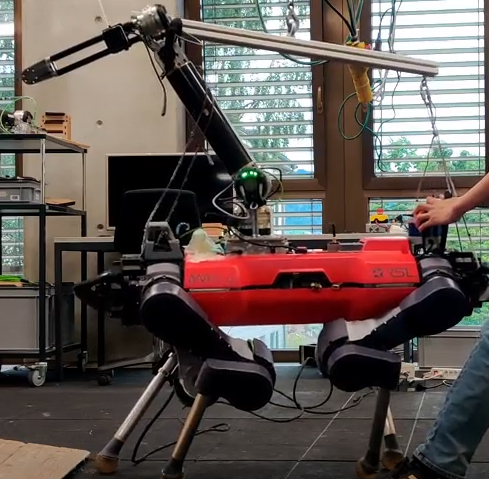}
    \vspace{-3pt}
    \label{subfig:ee_tracking1}}
    \vspace{3pt}
    \hfil
    \subfloat[]{\includegraphics[width=0.199\textwidth]{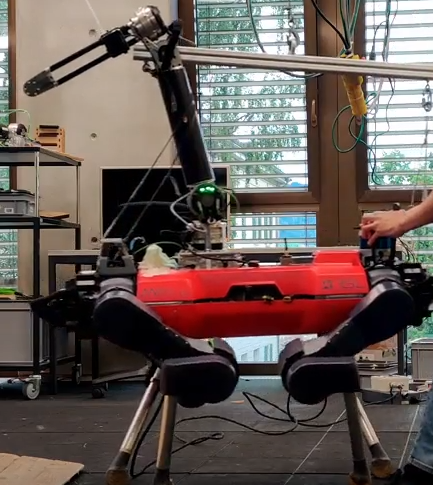}
    \vspace{-3pt}
    \label{subfig:ee_tracking1}}
    \vspace{3pt}
    \caption{Testing MPC's ability to track the desired end-effector position under base disturbances.}
    \vspace{-1pt}
    \label{fig:compare_student}
\end{figure}

We disturb base motion manually while ANYmal is in stance to evaluate \ac{MPC}'s ability to track the desired end-effector position {\color{Black}{under disturbances}}. This hardware experiment showed a 57\% reduction of the {\color{Black}{end-effector's}} position deviation {\color{Black}{compared to the base}} as we disturb the base in the $x$ direction. This result is based on our default controller setting. 
The performance for disturbance rejection can be further improved by adjusting the cost formulation of \ac{MPC} and the manipulator configuration.

{
\color{Black}{
\subsection{Response to External Push: Anticipated Leaning}
\begin{figure*}
    \centering
    \subfloat[]{\includegraphics[width=0.15\textwidth]{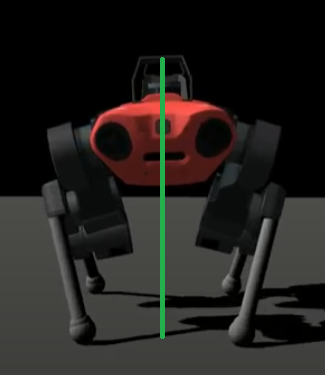}
    \label{subfig:lean1c}}
    \hfil
    \subfloat[]{\includegraphics[width=0.15\textwidth]{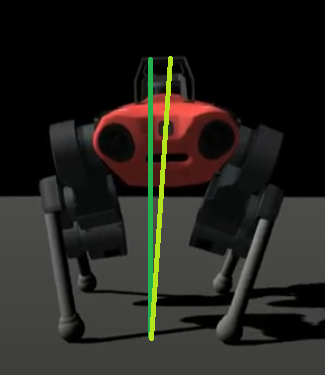}
    \label{subfig:lean2c}}
    \hfil
    \subfloat[]{\includegraphics[width=0.15\textwidth]{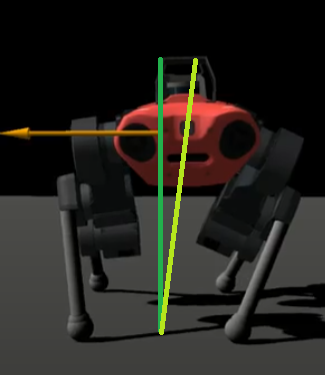}
    \label{subfig:lean3c}}
    \hfil
    \subfloat[]{\includegraphics[width=0.15\textwidth]{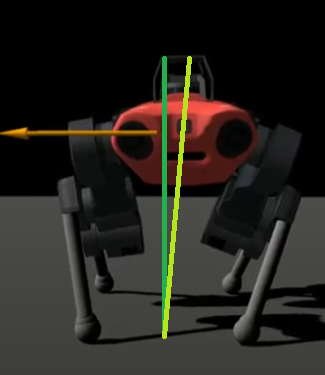}
    \label{subfig:lean4c}}
    \hfil
    \subfloat[]{\includegraphics[width=0.15\textwidth]{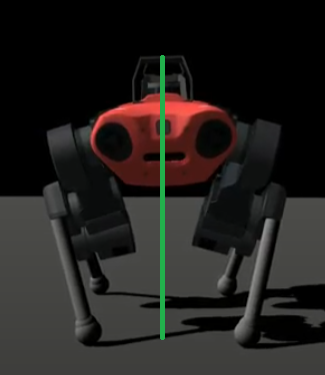}
    \label{subfig:lean5c}}
    \vspace{5pt}
    \caption{Our proposed controller uses the wrench prediction to leaning against an incoming force of \unit[100]{N} to maintain stability. (\ref{subfig:lean1c}) Upright stance when no external force is anticipated. (\ref{subfig:lean2c}) ANYmal leans right when it expects a force pulling from its left. (\ref{subfig:lean3c}) ANYmal stands stably when the external force is applied. (\ref{subfig:lean4c})-(\ref{subfig:lean5c}) ANYmal expects the pulling force to end, and returns to the upright posture.}
    \vspace{-10pt}
    \label{fig:lean}
\end{figure*}
This experiment tests how our proposed controller can utilize the external wrench prediction to gain more robustness. To this end, we simulate a lateral force as a step function of time to the robot's base. The policy observes this external force as part of its predicted wrench input. We quantify the robot's stability as a maximum force that it can tolerate before stepping. As the policy receives a prediction of the incoming force, it starts leaning against it before the actual push happens in order to increase its stability. Figure \ref{fig:lean} shows a qualitative response. This behavior is compared to policies trained with the same teacher, but with different observation input: (1) Reactive Controller: it only observes the current wrench estimation, (2) Naive Controller: it has no wrench observations. 

In each test, we set ANYmal to the same joint configuration and the same gait phase at the start of the episode and apply a force in $y$-direction that lasts for \unit[1]{s}. This experiment is deterministic for each policy, and student policies trained with five seeds are used for each setting to ensure reproducibility. We increase the force with the decimation of \unit[10]{N} to record the maximum force that the policy can stabilize without stepping. Since our control policy can leverage the force prediction and adjust its posture before experiencing the force, the robot could stabilize under a larger force of \unit[150]{N} without stepping to the side (\unit[170]{\%} of the reactive controller, and \unit[208]{\%} of the naive controller). The policy that does not observe the external wrench could potentially counteract the force using the proprioceptive history. Still, the transient response is not as prompt as the policy that directly observes the current wrench.


}
}

\begin{table}[]
\caption{\textsc{Comparison Between Using The Wrench Prediction Observation And Decoded Wrench}} \label{tab:decode}
\vspace{3pt}
\begin{tabular}{p{2.5cm}l|llll}
\multicolumn{2}{p{3.5cm}|}{Angular frequency $\omega$ / rad $s^{-1}$}                                              & 0       & 2     & 4     & 5     \\ \Xhline{2\arrayrulewidth}
\multicolumn{1}{p{1cm}|}{\multirow{2}{2.5cm}{Base angular velocity \newline error / (deg $s^{-1}$)}}                         & ours   & \textbf{0.234}  & \textbf{0.241} & \textbf{0.324} & \textbf{0.323} \\
\multicolumn{1}{p{2.5cm}|}{}                                                 & decode & 0.239  & 0.244 & 0.539 & 0.540 \\ \hline
\multicolumn{1}{p{2.5cm}|}{\multirow{2}{2.5cm}{Average time \newline before falling / $s$}} & ours   & -       & -     & \textbf{-}     & \textbf{4.1}   \\
\multicolumn{1}{p{2.5cm}|}{}                                                 & decode & -       & -     & 30    & 1.9  
\end{tabular}
\vspace{-3pt}
\end{table}

\subsection{Explicit Wrench Compensation}
As described in Sec.~\ref{sssec:student}, the \ac{RL} agent is trained to decode the external wrench from proprioceptive information. {\color{Black}{This decoded output could be fed back to the \ac{RL} actor and partially replace the external wrench prediction observation. In this simulated experiment, we compare ALMA's motion when using this explicit wrench compensation and discuss the advantage of using the wrench prediction instead of the decoded wrench.

Similarly to Experiment 2 in Sec.~\ref{ssec:ee_experiment}, we command the manipulator to move between the two sides of ALMA periodically to compare our proposed controller with the controller that replaces the wrench observation with the decoded wrench. The frequency that the arm moves cyclically is varied to evaluate the controllers. In this experiment, we use a \unit[3]{kg} end-effector and command the arm shoulder rotation joint angle $\theta$ to move according to $\theta_{\text{SH\_ROT}} = \frac{\pi}{2} \sin (\omega \tau)$, replacing the end-effector tracking cost with penalizing the deviation from commanded shoulder rotation for MPC. Tab.~\ref{tab:decode} indicates that using the decoded wrench to replace the MPC predictions results in a similar angular velocity tracking performance as our proposed controller when the arm moves slowly (when $\omega$ is below $2$ rad $s^{-1}$). But when the manipulator undertakes fast and dynamic motions, the wrench prediction 
reduces the angular velocity error of the base and the chance of falling due to the disturbance.
}}

{
\color{Black}{
\subsection{Dynamics Mismatch}
This experiment assesses the importance of training under unobserved wrench disturbance for handling the dynamics mismatch during deployment. To this end, we train another policy that is not exposed to the base-acceleration induced term of the unobserved wrench during the training and compare its performance against our policy in a setting similar to Sec.~\ref{sssec:compare_student} Experiment 1 with an additional, unobserved 1kg payload at the end-effector. 
Fig.~\ref{fig:dynamics_mismatch} shows a comparison of our performance indices for these two policies. While we observe a reduction in tracking error for every metric, we note a significant improvement with respect to the mean angular base velocity error, which translates to less undesired body motion in practice.

\begin{figure}
    \centering
    \includegraphics[width=0.36\textwidth]{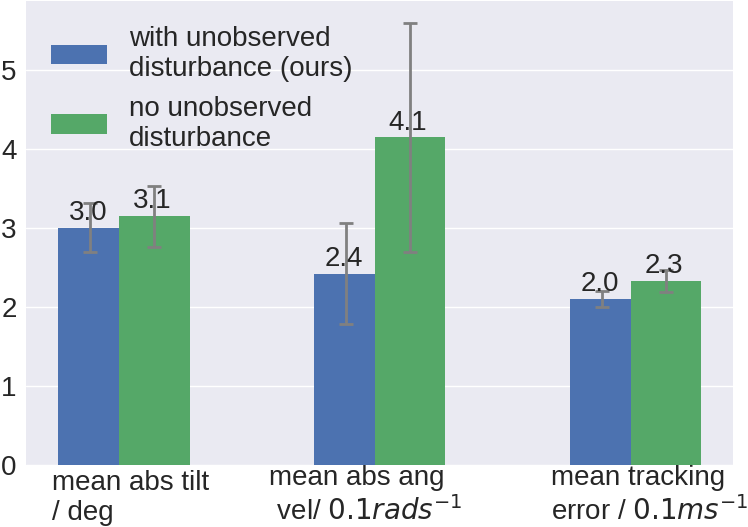}
    \vspace{7pt}
    \caption{Performance of the policies under dynamics mismatch.}
    \label{fig:dynamics_mismatch}
\end{figure}

}}

\vspace{-4pt}
\subsection{Dependency on Wrench Observations}
\begin{figure}[t]
    \centering
    \includegraphics[width=0.45\textwidth]{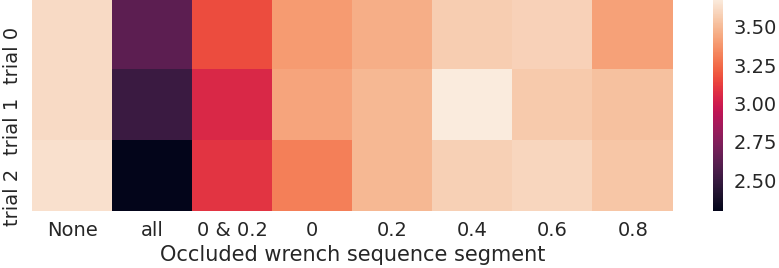}
    \vspace{3pt}
    \caption{Deterioration of the mean reward after occluding parts of the wrench observation.}
    \label{fig:occlusion}
    \vspace{-5pt}
\end{figure}

\vspace{-4pt}
We occlude segments of the wrench observation to evaluate the policy's dependency on each part of the observation. The occlusion is implemented by replacing each dimension of the observation segment with a normal distribution with the running mean and variance from the normalizer for that dimension. We rollout the policy for one hour in simulation time in the training environment. Fig.~\ref{fig:occlusion} shows the deterioration of the mean step rewards. For our wrench observation format, occluding the imminent wrench prediction (\unit[0.0]{s} and \unit[0.2]{s} in the future) leads to the most significant performance drop as expected. On average, the policy depends on the prediction at \unit[0.8]{s} more than those at \unit[0.4]{s} and \unit[0.6]{s}. We hypothesize that this is because the predictions at \unit[0.8]{s} are more informative than those at \unit[0.4]{s} and \unit[0.6]{s} as it evaluates the wrench trend further in the future.

%
%
\vspace{-4pt}
\section{Conclusion}
This work presented a pipeline to combine learning-based locomotion policy with model-based manipulation for legged mobile manipulators. We showed that the wrench sequence predictions from \ac{MPC} improve the base controller's performance on rejecting the external wrench. The combined controller adapts to a variety of manipulator configurations without re-training. Apart from manipulation, the policy is also suited for other applications such as load carrying and pulling. We demonstrated the controller on the robot hardware. To our knowledge, this is the first published result of legged mobile manipulation on rough terrain. 

Currently, our base policy responds to external wrench passively and has no control over it. An opportunity for future work is to allow the base policy to have partial control by tracking its desired wrench with manipulators. We hypothesize that this will enable more challenging maneuvers for the legged mobile manipulation systems.
\vspace{-4pt}

%
%

%
%
\bibliographystyle{IEEEtran}
\bibliography{sources} 
\end{document}